\documentclass[sigconf,screen,nonacm]{acmart}
\setkeys{acmart.cls}{balance=false}
\usepackage{amsmath}
\usepackage{booktabs}
\usepackage{colortbl}
\usepackage{array}
\usepackage{algorithm}
\usepackage{algpseudocode}

\setcopyright{none}
\AtEndPreamble{\hypersetup{colorlinks,linkcolor=ACMDarkBlue,citecolor=ACMDarkBlue,urlcolor=ACMDarkBlue}}

\definecolor{cedgreen}{RGB}{20,125,104}
\definecolor{cedblue}{RGB}{113,146,185}
\definecolor{cedorange}{RGB}{208,135,64}
\definecolor{cedrow}{RGB}{248,249,252}
\definecolor{cedwin}{RGB}{229,243,236}
\newcommand{\cedgain}[1]{\textcolor{cedgreen}{$\downarrow$#1\%}}
\definecolor{cedempty}{RGB}{214,221,230}
\newcount\cedcell
\newcommand{\ceilingcells}[2]{\makebox[26.4pt][l]{\cedcell=0\relax
  \loop\ifnum\cedcell<#2\relax
    \ifnum\cedcell<#1\relax\textcolor{cedgreen}{\rule{2.4pt}{6pt}}\else\textcolor{cedempty}{\rule{2.4pt}{6pt}}\fi
    \hspace{0.9pt}\advance\cedcell by 1\relax
  \repeat}}
\newcommand{\valuebar}[3]{\makebox[22pt][l]{\rlap{\textcolor{cedrow}{\rule{22pt}{3pt}}}\textcolor{#3}{\rule{#1pt}{3pt}}}\hspace{3pt}#2}

\title[CED-MoE: Reversible Service Tiers]{CED-MoE: Reversible Service Tiers\texorpdfstring{\\}{ }from Frozen Expert Routers}
\author{Linxiao Li}
\affiliation{\institution{AIGCode Co., Ltd.}\city{Shanghai}\country{China}}
\author{Weipeng Cao}
\affiliation{\institution{Guangdong Laboratory of Artificial Intelligence and Digital Economy (Shenzhen)}\city{Shenzhen}\country{China}}
\author{Jiongjiong Gu}
\affiliation{\department{Huawei Cloud}\institution{Huawei Technologies Co., Ltd.}\city{Shenzhen}\country{China}}
\author{Tingting Song}
\affiliation{\institution{Guangdong Laboratory of Artificial Intelligence and Digital Economy (Shenzhen)}\city{Shenzhen}\country{China}}
\author{Miao Liu}
\affiliation{\institution{Guangdong Laboratory of Artificial Intelligence and Digital Economy (Shenzhen)}\city{Shenzhen}\country{China}}
\author{Zhixiang Lu}
\affiliation{\institution{University of Liverpool}\city{Liverpool}\country{United Kingdom}}
\author{Fanyue Kong}
\affiliation{\department{2012 Lab}\institution{Huawei Technologies Co., Ltd.}\city{Shenzhen}\country{China}}
\author{Zhiying Li}
\affiliation{\institution{Xiaohongshu Inc.}\city{Shanghai}\country{China}}
\author{Minxian Xu}
\affiliation{\institution{Shenzhen Institutes of Advanced Technology, Chinese Academy of Sciences}\city{Shenzhen}\country{China}}
\author{Qiuwu Chen}
\affiliation{\institution{AIGCode Co., Ltd.}\city{Shanghai}\country{China}}
\email{chenqiuwu@aigcode.net}
\author{Zhong Ming}
\affiliation{\institution{Guangdong Laboratory of Artificial Intelligence and Digital Economy (Shenzhen)}\city{Shenzhen}\country{China}}
\email{mingz@szu.edu.cn}
\renewcommand{\shortauthors}{Linxiao Li et al.}

\makeatletter
\def\@mkauthors{%
  \global\setbox\mktitle@bx=\vbox{%
    \unvbox\mktitle@bx\par\medskip
    \centering
    {\fontsize{11}{15}\selectfont
      Linxiao Li\textsuperscript{1,*}\quad
      Weipeng Cao\textsuperscript{2,*}\quad
      Jiongjiong Gu\textsuperscript{3,*}\quad
      Tingting Song\textsuperscript{2}\quad
      Miao Liu\textsuperscript{2}\par
      Zhixiang Lu\textsuperscript{4}\quad
      Fanyue Kong\textsuperscript{5}\quad
      Zhiying Li\textsuperscript{6}\quad
      Minxian Xu\textsuperscript{7}\quad
      Qiuwu Chen\textsuperscript{1,\textdagger}\quad
      Zhong Ming\textsuperscript{2,\textdagger}\par}
    \vspace{5pt}
    {\fontsize{9}{11}\selectfont
      \textsuperscript{1}AIGCode Co., Ltd., Shanghai, China\par
      \textsuperscript{2}Guangdong Laboratory of Artificial Intelligence and Digital Economy (Shenzhen), Shenzhen, China\par
      \textsuperscript{3}Huawei Cloud, Huawei Technologies Co., Ltd., Shenzhen, China\par
      \textsuperscript{4}University of Liverpool, Liverpool, United Kingdom\par
      \textsuperscript{5}2012 Lab, Huawei Technologies Co., Ltd., Shenzhen, China\par
      \textsuperscript{6}Xiaohongshu Inc., Shanghai, China\par
      \textsuperscript{7}Shenzhen Institutes of Advanced Technology, Chinese Academy of Sciences, Shenzhen, China\par}
    \vspace{5pt}
    {\fontsize{8.5}{11}\selectfont
      \textsuperscript{*}Equal contribution.\quad
      \textsuperscript{\textdagger}Corresponding authors:
      \href{mailto:chenqiuwu@aigcode.net}{chenqiuwu@aigcode.net},
      \href{mailto:mingz@szu.edu.cn}{mingz@szu.edu.cn}.\par}
    \medskip
  }%
}
\makeatother

\begin{document}

\begin{abstract}
Web services that serve released Mixture-of-Experts (MoE) models must trade quality for cost as traffic changes, yet each router executes a fixed number of experts per token. Skipping experts in a frozen model offers this control, but prevailing skipping rules such as NAEE and ACE renormalize the retained gate weights. We find that this renormalization, which inflates the gain of the expert branch, raises perplexity more than six times as much as changing which experts are skipped. We introduce CED-MoE, which applies Calibrated Expert Deferral (CED). A prefix-relative router score decides which later experts execute, retained experts keep their native mixture coefficients, and one threshold calibrated on unlabeled router traces meets a requested average budget while capping the depth of every token. At matched compute on Qwen3-30B-A3B, CED lowers C4 perplexity by 17.5\% relative to both NAEE and ACE, and it wins all 18 calibrated comparisons against NAEE across three MoE families. On GSM8K, CED keeps 96.7\% of full-budget accuracy with 60.8\% fewer routed expert calls and leads NAEE by 16.9 points. A threshold calibrated once on C4 holds its budget from language modeling to arithmetic generation. Grouped execution converts the saving into up to 1.44$\times$ prefill throughput, and CED runs inside vLLM, where each service tier is one threshold that changes between batches without reloading the model.
\end{abstract}


\maketitle
\pagestyle{plain}
\hypersetup{pdfauthor={Linxiao Li, Weipeng Cao, Jiongjiong Gu, Tingting Song, Miao Liu, Zhixiang Lu, Fanyue Kong, Zhiying Li, Minxian Xu, Qiuwu Chen, Zhong Ming},pdfkeywords={Mixture-of-experts, LLM serving, service tiers, budget calibration, Web services}}

\section{Introduction}

Retrieval-augmented assistants and Web agents place large language models behind search, question answering, and tool use \citep{lewis2020rag,yao2023react}. These Web services answer requests under different latency and cost targets. Web servers have long met such targets under overload by serving lighter content instead of dropping requests \citep{abdelzaher1999web,klein2014brownout}. Serving systems optimize batching and memory \citep{yu2022orca,kwon2023vllm,zheng2024sglang,agrawal2024sarathi,baton2025www}, and SCOOT tunes inference engines for service-level objectives \citep{cheng2025scoot}. Released MoE models such as Mixtral, DeepSeek-V2, and Qwen3 bring sparse expert computation to deployment \citep{jiang2024mixtral,dai2024deepseekmoe,dai2024deepseekv2,qwen2025qwen3}. Their routers choose expert identities for each token, yet the native number of executed experts remains fixed. CED exposes expert depth as a calibrated budget that serving infrastructure can set per service tier and reverse at any time.

\begin{figure*}[t]
\centering
\includegraphics[width=\textwidth]{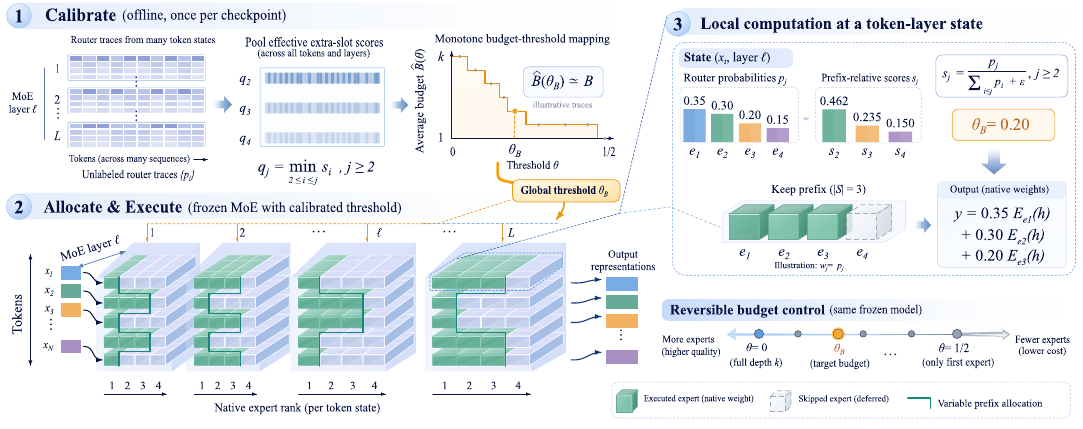}
\caption{CED-MoE turns native expert rankings into a calibrated, reversible compute budget. Unlabeled traces calibrate one shared threshold from effective prefix scores. Teal prefixes execute with native coefficients, while pale suffixes defer. The grid and budget curve use illustrative top-4 states with $w_j=p_j$. In the enlarged state, scores 0.462, 0.235, and 0.150 under $\theta_B=0.20$ retain three experts.}
\Description{A calibration lane pools effective prefix scores and maps a target budget to one threshold. Four exploded MoE-layer grids show five illustrative token states with one or three of four experts executing, averaging two experts. The empirical budget curve is computed from the same states. The enlarged state retains native coefficients 0.35, 0.30, and 0.20. A reversible threshold control spans full depth to the first expert.}
\label{fig:overview}
\end{figure*}

Expert computation can be adapted through learned routing components that require additional training \citep{zeng2024adamoe,jin2024moepp,huang2024harder,yue2024adak,guo2024dynmoe}. On frozen models, layer-level allocators set how many experts each layer executes \citep{chittyvenkata2025lexi,liu2026allocmoe}, and skipping rules decide which calls each token omits. NAEE (Not All Experts are Equal) \citep{lu2024experts}, the closest such rule, skips a later expert by its weight ratio to the leading expert. Its released implementation and its tail-mass form for longer lists, which is top-$p$ selection \citep{huang2024harder}, both renormalize the retained weights.

Router weights both rank experts and scale their contributions. CED separates these two roles. It scores a later expert by its share of the prefix's routing mass, then preserves the native mixture coefficient when the expert executes. Preserving coefficients confines the local output change to the omitted suffix, whereas renormalization also amplifies every retained contribution. The scaling role matters more. At matched depth, keeping native coefficients lowers perplexity more than six times as much as changing which experts are skipped.

Unlabeled calibration translates a requested average depth into one shared threshold, and each token executes the longest native prefix whose extra-slot scores exceed it. Full depth restores the native model exactly.

At target $B=3.2$, a C4-calibrated CED threshold executes 3.10 to 3.25 experts from held-out language modeling to GSM8K decoding, where absolute-probability thresholding drops to 1.82. CED lowers perplexity relative to NAEE in all 18 family, corpus, and budget comparisons, by up to 75.3\%, and beats ACE~\citep{xu2026ace} by 17.5\%. On full GSM8K, CED exceeds NAEE by 16.9 accuracy points while executing 8.9\% fewer expert calls.

Our contributions are:
\begin{itemize}
    \item \textbf{A calibrated, reversible compute control.} A router-only prefix policy couples one shared budget threshold with native mixture coefficients. The threshold also yields a per-token depth ceiling and an explicit budget-transfer identity.
    \item \textbf{Skipping without renormalization.} Native coefficients matter more than which experts are skipped. They account for most of CED's gain over renormalized skipping in all three MoE families, and CED wins all 18 calibrated comparisons against NAEE.
    \item \textbf{A budget that survives deployment.} On Qwen3, one C4 threshold transfers within 3.1\% of its target across language modeling and reasoning, preserves 96.7\% of full GSM8K accuracy, and reaches 1.44$\times$ prefill throughput. The same policy runs inside vLLM.
\end{itemize}

\section{Calibrated Expert Deferral}

\subsection{Native Experts as an Ordered Cascade}

For token $t$ at layer $l$, the frozen router selects a native list of $k$ experts from $N$ candidates. We order this list by its selected router scores:
\begin{equation}
    (e_1,\ldots,e_k)=\operatorname{NativeTopK}(R_l(h_{l,t})).
\end{equation}
The selection preserves the checkpoint's routing constraints, including grouped selection in DeepSeek-V2. Let $p_j$ be the selected router probability before top-$k$ normalization and $w_j$ its native mixture coefficient. Qwen3 and Mixtral normalize selected probabilities. DeepSeek-V2 applies its native routed scaling factor. With $E_e$ denoting expert $e$, the routed component is
\begin{equation}
    y^{\mathrm{native}}_{l,t}=\sum_{j=1}^{k}w_jE_{e_j}(h_{l,t}).
\end{equation}
CED controls the depth of this native expert list.

\subsection{Prefix Deferral with Native Weights}

CED retains the first expert and selects a prefix. With $z_{l,t,j}$ indicating whether slot $j$ executes,
\begin{equation}
    z_{l,t,1}=1,\qquad z_{l,t,j}\le z_{l,t,j-1}\quad(j\ge2),
\end{equation}
\begin{equation}
    y^{\mathrm{CED}}_{l,t}=\sum_{j=1}^{k}z_{l,t,j}w_jE_{e_j}(h_{l,t}).
\end{equation}
Every retained expert keeps its native coefficient. Skipping a suffix removes its contribution without rescaling the experts that remain.

The prefix-relative score for slot $j$ is
\begin{equation}
    s_{l,t,j}=\frac{p_j}{\sum_{i=1}^{j}p_i+\epsilon},
    \qquad j=2,\ldots,k,
\end{equation}
with $\epsilon=10^{-9}$. The score measures the current expert's share of its prefix mass. A dominant leading expert drives later scores toward zero, whereas a flat routing distribution keeps them at their largest values, so one threshold gives confident tokens short prefixes and uncertain tokens long ones. Apart from the stabilizer, it is invariant to a common positive rescaling of router probabilities. A shared threshold $\theta_B$ determines the retained depth:
\begin{equation}
\begin{aligned}
    m_{l,t}(\theta_B)
    &=\max\{m\in\{1,\ldots,k\}:\\
    &\qquad s_{l,t,j}>\theta_B,\ j=2,\ldots,m\}.
\end{aligned}
\end{equation}
A slot executes when it and every earlier extra slot exceed the threshold. Define the effective prefix score
\begin{equation}
    q_{l,t,j}=\min_{2\le i\le j}s_{l,t,i},\qquad j\ge2.
    \label{eq:effective-score}
\end{equation}
The retained depth then has the threshold form
\begin{equation}
    m_{l,t}(\theta_B)=1+\sum_{j=2}^{k}
    \mathbb{I}[q_{l,t,j}>\theta_B].
    \label{eq:threshold-depth}
\end{equation}

For two illustrative states, state A stops after its first expert because $s_2\approx0.176<0.20$. State B, shown in Figure~\ref{fig:overview}, retains three experts, with extra-slot scores 0.462, 0.235, and 0.150. The shared threshold thus meets the target mean depth of two while keeping every retained coefficient.

\paragraph{Relation to ratio skipping.}
For $k=2$ and $\epsilon=0$, let $r=p_2/p_1$. Then $s_2=r/(1+r)$, so CED and NAEE's top-2 score induce the same ordering \citep{lu2024experts}, with threshold mapping $\theta=\beta/(1+\beta)$. For longer lists, CED compares a slot with its own prefix. NAEE's top-$k$ extension thresholds the entire removable tail relative to the native top-$k$ mass.

\subsection{Calibrating an Average Expert Budget}

Given $C=|\mathcal C|$ unlabeled token-layer records, the empirical average depth is
\begin{equation}
    \hat B(\theta)=\frac{1}{|\mathcal C|}
    \sum_{(l,t)\in\mathcal C}m_{l,t}(\theta).
\end{equation}
CED rounds the requested extra-call count to $M=\mathrm{round}(C(B-1))$ and selects the nearest attainable count:
\begin{equation}
    \theta_B\in\operatorname*{arg\,min}_{\theta}
    |C(\hat B(\theta)-1)-M|.
    \label{eq:calibration}
\end{equation}
The implementation pools the effective extra-slot scores and uses empirical order statistics to retain approximately $|\mathcal C|(B-1)$ calls. At the endpoints, $B=1$ executes only the first slot and $B=k$ executes the full native list. Interior budgets use the fitted threshold.

One threshold is shared across layers and extra slots. One set of router traces supports every requested budget without further model passes. Our Qwen profiles request $B=2.0,3.2,4.5$, DeepSeek profiles request $B=2.0,3.0,4.0$, and Mixtral profiles request $B=1.2,1.5,1.8$.

\begin{algorithm}[t]
\caption{Calibrated expert deferral}
\label{alg:ced}
\begin{algorithmic}[1]
\Require Frozen router $R_l$, experts $E_e$, traces $\mathcal C$, budget $B$
\State Compute effective prefix scores $q$ on $\mathcal C$
\State Set desired extra-call count $M\gets\mathrm{round}(|\mathcal C|(B-1))$
\State Find score order statistics bracketing count $M$
\State Choose $\theta_B$ by Eq.~\eqref{eq:calibration}, keeping ties intact
\For{each inference batch at layer $l$}
  \State Read native tuples $(e_j,p_j,w_j)$
  \State Compute prefix scores $s_{l,t,j}$
  \State Set depth to $B$ at endpoints $B\in\{1,k\}$
  \State Otherwise use Eq.~\eqref{eq:threshold-depth}
  \State Group retained token-expert pairs by expert index
  \State Execute each nonempty expert batch
  \State Accumulate outputs with native coefficients $w_j$
\EndFor
\end{algorithmic}
\end{algorithm}

\paragraph{Monotone control and full retention.}
Raising $\theta$ can remove only suffix slots, so $\hat B(\theta)$ is non-increasing. At full retention, $z_{l,t,j}=1$ for every slot and the native and CED expressions coincide. Algorithm~\ref{alg:ced} gives the calibration and execution procedure.

\paragraph{Calibration cost.}
Computing the $C(k-1)$ scores takes $O(Ck)$ work, every requested budget reuses them through order-statistic selection, and the online decision requires $O(k)$ scalar operations per routed token-layer state.

\paragraph{A per-token compute ceiling.}
Descending probabilities give $p_j\le P_j/j$, where $P_j=\sum_{i\le j}p_i$. Consequently, $s_j\le 1/j$. A positive threshold therefore imposes an upper bound on every routed depth:
\begin{equation}
 m_{l,t}(\theta)\le
 \min\!\left\{k,\max\!\left\{1,\left\lceil\frac{1}{\theta}\right\rceil-1\right\}\right\}.
 \label{eq:depth-ceiling}
\end{equation}
The mandatory first expert accounts for the lower bound of one. For Qwen's calibrated $B=3.2$ profile, $\theta_B\approx0.20325$ limits each token-layer state to at most four of the eight native experts. A serving system can determine this maximum assignment count directly from the threshold before observing a request's router scores.

\paragraph{From score stability to budget stability.}
The threshold also gives an explicit description of transfer. Let $F_{\mathcal C}$ be the empirical CDF of all effective extra-slot scores in the calibration records. Equation~\eqref{eq:threshold-depth} gives
\begin{equation}
 \hat B(\theta)=1+(k-1)\bigl(1-F_{\mathcal C}(\theta)\bigr).
\end{equation}
For a new workload, let $F_{\mathcal D}$ be the corresponding score CDF along its executed trajectories and $B_{\mathcal D}$ its achieved depth. At the frozen threshold,
\begin{equation}
 B_{\mathcal D}-\hat B=(k-1)
 \bigl(F_{\mathcal C}(\theta_B)-F_{\mathcal D}(\theta_B)\bigr).
 \label{eq:budget-transfer}
\end{equation}
Thus the budget error is bounded by calibration error plus the change in score distribution:
\begin{equation}
 |B_{\mathcal D}-B|\le |\hat B-B|
 +(k-1)\|F_{\mathcal D}-F_{\mathcal C}\|_\infty.
\end{equation}
This identity separates the requested operating point from the workload's routing statistics.

\subsection{Preserving the Retained Contribution}

For fixed hidden state $h_{l,t}$ and depth $m$, preserving the native coefficients yields
\begin{equation}
    y^{\mathrm{native}}_{l,t}-y^{\mathrm{CED}}_{l,t}
    =\sum_{j=m+1}^{k}w_jE_{e_j}(h_{l,t}).
    \label{eq:suffix-deviation}
\end{equation}
The change is exactly the omitted suffix contribution. Assume each expert output has Euclidean norm at most $M$. Then
\begin{equation}
    \|y^{\mathrm{native}}_{l,t}-y^{\mathrm{CED}}_{l,t}\|_2
    \le M\sum_{j=m+1}^{k}w_j.
    \label{eq:mass-bound}
\end{equation}
Renormalization also changes the retained terms. We isolate this effect by redistributing the original selected mass $W=\sum_i w_i$ over the retained prefix:
\begin{equation}
    \widetilde w_j=z_jw_j\frac{W}{\sum_i z_iw_i}.
    \label{eq:renormalization}
\end{equation}
For Qwen3 and Mixtral, $W=1$. For DeepSeek-V2, this convention preserves the checkpoint's native routed mass and scaling.

Write the omitted fraction of native mass as $\alpha=W^{-1}\sum_{j>m}w_j$. Renormalization multiplies every retained contribution by $1/(1-\alpha)$, inflating the gain of the expert branch as more experts are skipped. Its local deviation can be written as
\begin{equation}
    y^{\mathrm{native}}-y^{\mathrm{renorm}}
    =\sum_{j>m}w_jE_{e_j}
    -\frac{\alpha}{1-\alpha}\sum_{j\le m}w_jE_{e_j}.
\end{equation}
Under the same expert-norm bound, the two terms give
\begin{equation}
    \|y^{\mathrm{native}}-y^{\mathrm{renorm}}\|_2\le 2MW\alpha,
\end{equation}
whereas Eq.~\eqref{eq:mass-bound} gives $MW\alpha$ for native-weight deferral. Both worst-case bounds are attained when retained and omitted expert outputs point in opposite directions with norm $M$. Preserving the coefficients removes the additional perturbation from amplifying retained experts.

\begin{proposition}[Calibrated control]\label{prop:control}
For any threshold $\theta>0$, CED executes a native prefix that contains the first expert, and every token-layer state executes at most $\min\{k,\max\{1,\lceil1/\theta\rceil-1\}\}$ experts. On a workload with effective-score CDF $F_{\mathcal D}$, the achieved depth obeys $|B_{\mathcal D}-B|\le|\hat B-B|+(k-1)\|F_{\mathcal D}-F_{\mathcal C}\|_\infty$. With expert outputs bounded by $M$, native coefficients limit the local output change to $MW\alpha$, half of the renormalized worst case $2MW\alpha$.
\end{proposition}
\begin{proof}[Proof sketch]
Prefix closure follows from the running minimum in Eq.~\eqref{eq:effective-score}, and the ceiling from $s_j\le1/j$. The transfer bound follows from Eq.~\eqref{eq:budget-transfer}, and the deviation bounds from Eq.~\eqref{eq:suffix-deviation} and the renormalized decomposition.
\end{proof}

\subsection{Grouped Expert Execution}

For $T$ routed tokens, native inference creates $Tk$ token-expert assignments. CED creates $T\bar m$, where $\bar m$ is achieved average depth. Descending router scores make the threshold mask prefix-closed directly. We gather retained assignments by expert, execute their MLPs with grouped matrix multiplications, and scatter the weighted outputs back to tokens.

The routed expert-call reduction is
\begin{equation}
    \mathrm{Saving}=1-\frac{\bar m}{k}.
    \label{eq:saving}
\end{equation}
For Qwen's balanced target $B=3.2$, nominal reduction is 60\%. The held-out C4 workload executes 3.15 experts per routed token-layer, a 60.6\% reduction.

\section{Experiments}

\subsection{Evaluation Setup}

We evaluate Qwen3-30B-A3B, Mixtral-8x7B-v0.1, and DeepSeek-V2, whose native routed depths are eight, two, and six. Model weights remain frozen throughout.

\paragraph{Calibration and language modeling.}
Each model supplies full-budget router traces from 128 distinct C4 training documents \citep{raffel2020t5}. We select one deterministic 512-token input window per document and fit all budget thresholds to these traces. Frozen thresholds are evaluated on 128 held-out C4 validation documents and 128 non-overlapping WikiText-2 test blocks \citep{merity2016wikitext}. Both are Web corpora, drawn from Common Crawl and Wikipedia. Paired bootstrap intervals resample documents or blocks 5,000 times \citep{efron1993bootstrap}.

\paragraph{Baselines.}
Full-budget grouped execution provides the common reference. NAEE spans ratio and top-$p$ skipping. Its top-$k$ tail rule is top-$p$ selection \citep{huang2024harder} on the normalized native list, and we fit both shared and layer-specific NAEE thresholds to each requested budget. On Mixtral, we also evaluate NAEE's original per-layer median ratio rule, and on Qwen the released ACE code \citep{xu2026ace}, calibrated on the same traces. Random adjacent-depth prefixes, static prefixes, and absolute-probability thresholds serve as allocation controls. A count-matched comparison refits thresholds on the unlabeled evaluation inputs to compare selection and normalization at equal achieved depth.

\paragraph{Reasoning and runtime.}
GSM8K uses all 1,319 test questions, five fixed training demonstrations with worked solutions \citep{cobbe2021gsm8k,wei2022cot}, Qwen's native chat template with thinking disabled, greedy decoding, and a 2,048-token generation limit. A common numerical-answer matcher extracts the final number, and Appendix~\ref{app:generation} also reports strict delimiter-based matching. The grouped policies use batch size 16, and all measurements use BF16 weights. Appendix~\ref{app:implementation} specifies the execution environments.

\subsection{Reasoning with 60.8\% Fewer Expert Calls}

\begin{figure}[t]
\centering
\includegraphics[width=\columnwidth]{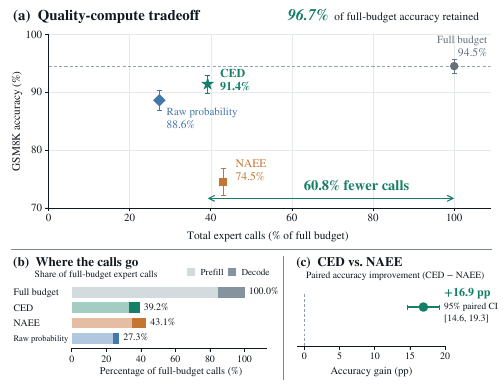}
\caption{CED keeps 96.7\% of full-budget GSM8K accuracy with 60.8\% fewer routed expert calls. (a) Accuracy on all 1,319 questions versus total calls. (b) Prefill (light) and decode (dark) calls as shares of the full-budget total. (c) Paired gain over NAEE. Intervals are 95\% question-bootstrap intervals. Thresholds are frozen after C4 calibration at $B=3.2$.}
\Description{Three panels compare GSM8K accuracy and routed work. The main scatter shows CED at 91.4 percent accuracy and 39.2 percent of full-budget calls, NAEE at 74.5 and 43.1 percent, raw probability at 88.6 and 27.3 percent, and full budget at 94.5 percent accuracy. Stacked bars split each total into prefill and valid decode calls. The paired CED-minus-NAEE gain is 16.9 percentage points with a 95 percent interval of 14.6 to 19.3.}
\label{fig:gsm-quality}
\end{figure}

CED preserves 96.7\% of full-budget accuracy while removing 60.8\% of routed expert calls on GSM8K. It reaches 91.43\% against 94.54\% for full-budget grouped execution. The C4-calibrated threshold executes 3.10 experts during prefill and 3.25 during decode. Figure~\ref{fig:gsm-quality} compares accuracy with total routed calls, counting the generated trajectory of every question.

NAEE reaches only 74.53\% under the same calibration target, while CED executes 8.9\% fewer expert calls. CED's 16.9-point advantage has a paired 95\% interval of [14.6, 19.3] points, and CED answers 252 questions that NAEE misses against 29 in the other direction. Absolute-probability thresholding undershoots the requested budget, with achieved depth falling to 1.82 during decoding, and still trails CED by 2.8 points with exact McNemar $p<0.001$ \citep{mcnemar1947note}. NAEE also destabilizes generation, and its strict delimiter-based accuracy collapses to 59.59\% against 91.36\% for CED.

\subsection{Quality across Calibrated Budgets}

CED supplies a range of operating points between one expert and the native depth. At the lowest Qwen budget, per-layer NAEE executes more experts than CED, 2.05 against 1.97, and its C4 perplexity explodes to 69.19, while CED reaches 25.79, a 62.7\% reduction. At $B=4.5$, CED stays within 11.4\% of full-budget perplexity while removing 44.4\% of routed calls.

Score-based allocation also beats score-free prefix allocation. Random adjacent-depth prefixes at the full target depth 3.20 reach C4 perplexity 19.36 to 19.38. CED reaches 18.90 while executing fewer experts, 3.15 per routed token-layer.

\begin{figure}[t]
\centering
\includegraphics[width=\columnwidth]{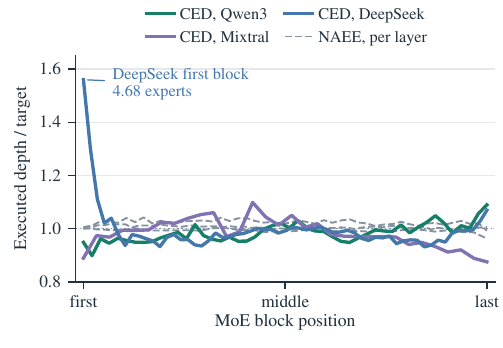}
\caption{One shared threshold allocates depth across layers. Lines show executed depth per MoE block relative to the target at the middle C4 profiles. CED concentrates calls where routers spread mass, while layer-specific NAEE thresholds hold every block near the target.}
\Description{Per-block executed depth divided by the target for CED on Qwen3, Mixtral, and DeepSeek, with layer-specific NAEE as flat dashed lines. DeepSeek's first MoE block reaches 4.68 experts under CED, about 1.56 times its target.}
\label{fig:layer-depth}
\end{figure}

One shared threshold also distributes depth across layers without any per-layer tuning (Figure~\ref{fig:layer-depth}). DeepSeek's first MoE block receives 1.7 times the depth of its lightest block, while layer-specific NAEE quantiles hold every block near the target and still trail CED at every budget.

The ordering carries to WikiText-2 with the C4 thresholds frozen. CED reaches perplexity 14.73 at 3.11 experts, compared with 19.64 for per-layer NAEE at 3.25 experts and 19.11 for ACE at 3.24. Every CED profile executes fewer experts and reaches lower perplexity than the matching NAEE profile on both corpora (Table~\ref{tab:quality-qwen}).

Mixtral isolates retained-weight scaling because its top-2 CED score reparameterizes NAEE's ratio. At target $B=1.5$, CED reaches C4 and WikiText-2 perplexities 7.69 and 5.86. Renormalizing CED reproduces shared-threshold NAEE exactly, at 7.78 and 6.00, and NAEE's original median rule reaches 7.79 and 6.01. Native coefficients alone therefore account for the entire Mixtral gain. At target $B=1.8$, CED removes 10.8\% of C4 expert calls with perplexity 7.56 against 7.52 at full depth.

\begin{figure*}[t]
\centering
\includegraphics[width=\textwidth]{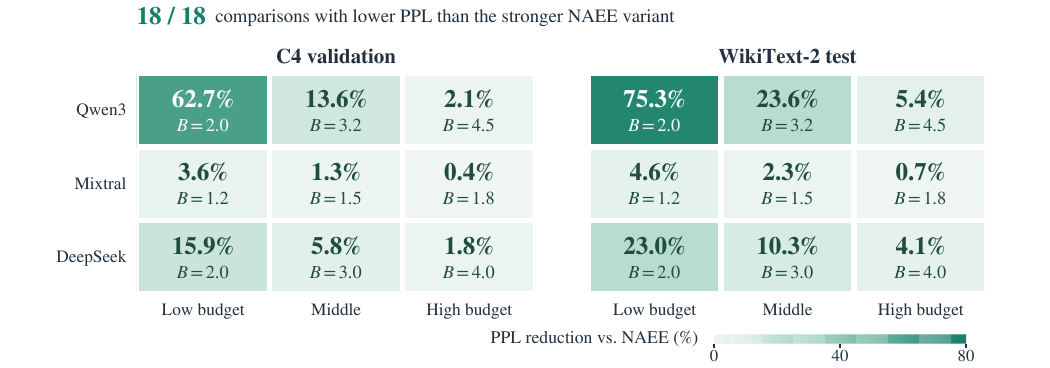}
\caption{CED lowers perplexity in all 18 family, corpus, and budget comparisons. Each cell reports the PPL reduction relative to the better of shared-threshold and per-layer NAEE at the same calibration target $B$, and all 18 paired document-bootstrap 95\% intervals exclude parity. Thresholds remain frozen after C4 training calibration. Executed depths and absolute PPL values appear in Appendix~\ref{app:quality}.}
\Description{Two heatmaps show eighteen positive perplexity reductions over the better NAEE threshold variant, across Qwen3, Mixtral, and DeepSeek at three calibrated budget levels on C4 and WikiText-2. Every cell includes its numerical target budget.}
\label{fig:cross-family-gains}
\end{figure*}

DeepSeek sustains quality with a longer native expert list. At target $B=2$, CED executes 1.99 experts and reaches C4 perplexity 8.29, compared with 9.85 for per-layer NAEE at 2.02 experts. CED removes 66.9\% of routed expert calls while keeping perplexity within 6.1\% of full-budget execution, and at $B=4$ it stays within 1.0\% with 33.6\% fewer calls. Figure~\ref{fig:cross-family-gains} extends the comparison to every calibrated budget. CED has lower perplexity in all 18 cases, even against the better NAEE threshold variant at each point, and every paired 95\% interval excludes parity. The margin widens monotonically as the budget tightens in every family and on both corpora, reaching 62.7\% on Qwen C4 and 23.0\% on DeepSeek WikiText-2 at the lowest budget.

\begin{table}[t]
\caption{Qwen achieved expert depth at target $B=3.2$. The upper block uses one C4 calibration, with layer-specific NAEE thresholds. The lower block calibrates on WikiText-2 validation traces collected at full depth or with only the first expert executing, with a shared NAEE threshold. The last row gives the largest deviation from the target.}
\label{tab:budget-transfer}
\centering
\small
\setlength{\tabcolsep}{4pt}
\begin{tabular}{@{}lrr>{\columncolor{cedwin}}r@{}}
\toprule
Workload & Raw $p_j$ & NAEE & \textbf{CED} \\
\midrule
C4 validation & 3.095 & 3.276 & 3.149 \\
WikiText-2 test & 3.077 & 3.252 & 3.115 \\
GSM8K prefill & 2.261 & 3.289 & 3.102 \\
GSM8K decode & 1.817 & 3.444 & 3.249 \\
\midrule
Full-depth, C4 & 3.142 & 3.362 & 3.226 \\
Full-depth, Wiki & 3.125 & 3.337 & 3.190 \\
First-slot, C4 & 4.265 & 3.333 & 3.285 \\
First-slot, Wiki & 4.242 & 3.308 & 3.247 \\
\midrule
Largest error$\downarrow$ & 43.22\% & 7.62\% & \textbf{3.05\%} \\
\bottomrule
\end{tabular}
\end{table}

\subsection{Native Coefficients Outweigh Selection}

The matched-depth experiment crosses two prefix scores with two weight treatments. All four policies and ACE execute approximately 3.15 experts on the same 128 C4 documents. With a shared threshold for each policy, CED with native coefficients reaches perplexity 18.90, compared with 22.89 for renormalized NAEE and 22.90 for ACE, a 17.5\% perplexity reduction at matched computation.

\begin{table}[t]
\caption{CED is best at matched compute, and native coefficients drive the gain. Count-matched Qwen C4 comparisons, all within 0.005 of CED's depth 3.1491. CED gain is its PPL reduction against each row. NAEE uses shared or per-layer thresholds, and CED and ACE keep one shared threshold.}
\label{tab:matched-depth}
\centering
\begin{tabular}{@{}lllrrr@{}}
\toprule
Scope & Score & Weights & Depth & PPL$\downarrow$ & CED gain \\
\midrule
\rowcolor{cedwin}
Shared & \textbf{CED} & Native & 3.1491 & \textbf{18.90} & \\
 & NAEE & Renorm. & 3.1497 & 22.89 & \cedgain{17.5} \\
 & NAEE & Native & 3.1520 & 19.35 & \cedgain{2.4} \\
 & CED & Renorm. & 3.1463 & 22.32 & \cedgain{15.3} \\
 & ACE & Renorm. & 3.1465 & 22.90 & \cedgain{17.5} \\
\midrule
Per-layer & NAEE & Renorm. & 3.1478 & 23.19 & \cedgain{18.5} \\
 & NAEE & Native & 3.1525 & 19.47 & \cedgain{2.9} \\
\bottomrule
\end{tabular}
\end{table}

\begin{figure}[t]
\centering
\includegraphics[width=\columnwidth]{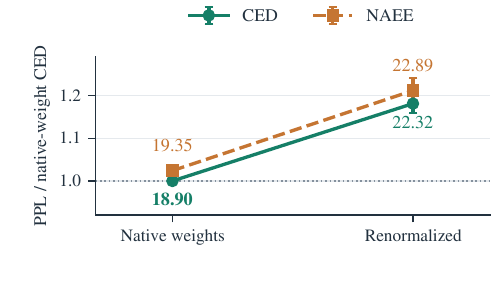}
\caption{Preserving native coefficients delivers the larger quality gain under both selection rules. Each line moves one selection rule from native to renormalized weights at a matched depth of about 3.15 experts on Qwen C4. The vertical axis is PPL relative to native-weight CED, labels give absolute PPL, and bars are paired document-bootstrap 95\% intervals.}
\Description{A two-factor plot compares native and renormalized weights under CED and NAEE selection. Both lines rise sharply with renormalization, while the CED line remains below NAEE.}
\label{fig:weight-rule}
\end{figure}

Preserving NAEE's coefficients lowers its perplexity by 15.5\% to 19.35, and renormalizing CED raises its perplexity to 22.32 (Figure~\ref{fig:weight-rule}). Native coefficients therefore account for most of the gap, and prefix-relative selection adds a further 2.4\% reduction, with a paired NAEE-to-CED ratio of 1.024 and 95\% interval [1.020, 1.029]. Layer-specific NAEE thresholds reproduce this ordering (Table~\ref{tab:matched-depth}). On DeepSeek, renormalizing CED likewise raises its C4 perplexity from 7.98 to 8.45, next to NAEE's 8.47. ACE renormalizes as well and, despite its parameter-based contribution estimates, lands next to renormalized NAEE at 22.90. Frozen-model skipping should therefore keep native coefficients, whichever of these rules selects the skipped experts.

\subsection{A Shared Threshold Transfers the Compute Budget}

A frozen C4 threshold holds CED depth between 3.10 and 3.25 across C4, WikiText-2, and the prefill and decode phases of GSM8K. Every value lies within 3.1\% of the requested depth 3.2 (upper block of Table~\ref{tab:budget-transfer}). Absolute-probability thresholds track the target on C4 and WikiText-2, then collapse to 2.26 experts during arithmetic prefill and 1.82 during decoding, losing 43\% of the requested budget. Prefix-relative normalization sustains the requested execution level as the generated trajectory changes. A common rescaling of router probabilities leaves every prefix-relative score unchanged, so the threshold keeps its meaning when a workload shifts overall router confidence.

\begin{figure}[t]
\centering
\includegraphics[width=\columnwidth]{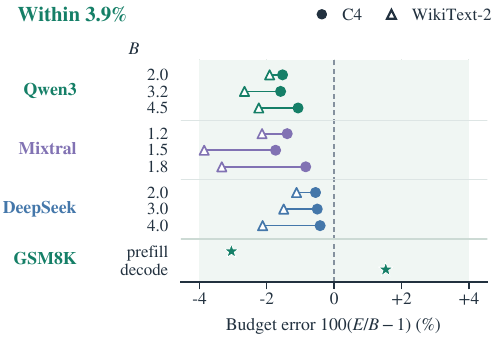}
\caption{Every calibrated CED profile lands within 3.9\% of its requested budget. Budget error is $100(E/B-1)$ for achieved depth $E$. Profiles are grouped by family with filled C4 and open WikiText-2 markers, stars show the Qwen GSM8K phases at $B=3.2$, and shading marks $\pm4\%$.}
\Description{Budget errors of eighteen CED language-model profiles grouped by Mixtral, Qwen3, and DeepSeek, plus two Qwen GSM8K phases. All points lie between minus 3.9 percent and plus 1.6 percent.}
\label{fig:budget-fidelity}
\end{figure}

The same fidelity holds across families. On held-out C4 and WikiText-2, all nine calibrated CED profiles of Qwen3, Mixtral, and DeepSeek execute within 3.9\% of their requested budgets and never exceed them (Figure~\ref{fig:budget-fidelity}). Renormalized skipping overshoots instead, and on Qwen every renormalized policy executes more experts than requested on both corpora, up to 3.28 at target 3.2 (Table~\ref{tab:quality-qwen}). A requested CED budget is therefore a reliable operating point, and operators can provision expert compute for a tier from its target before traffic arrives.

The budget also survives a change in how calibration traces are collected (lower block of Table~\ref{tab:budget-transfer}). Traces logged at full depth or with only the first expert executing both keep CED within 3\% of the target, while absolute-probability calibration shifts by more than one expert. A service can therefore calibrate CED from its own logged traffic, even traffic served at the cheapest depth, without a separate full-depth pass. Absolute-probability thresholds need full-depth traces to land near the same target.

\begin{figure*}[t]
\centering
\includegraphics[width=\textwidth]{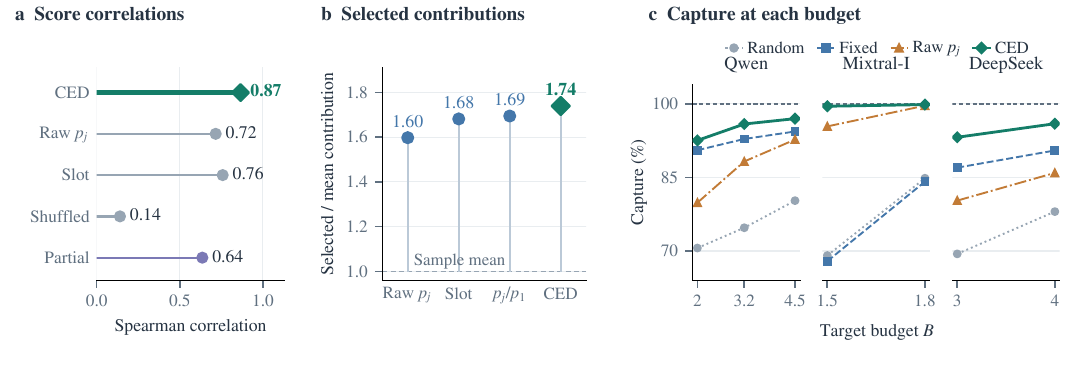}
\caption{Native router scores identify local expert contributions. (a) Qwen score correlations, including controls for raw probability, layer, and slot. (b) Mean contribution among selected slots divided by the sample mean, where $p_j/p_1$ scores a slot against the leading expert. (c) Contribution capture at each evaluated budget, as a percentage of a matched-size contribution-ranked prefix reference. Contribution is the local output change in Eq.~\eqref{eq:local-damage}, and Mixtral-I denotes Mixtral-8x7B-Instruct.}
\Description{Three panels show correlations between router scores and local prefix contributions, concentration of large contributions among selected slots, and captured contribution across Qwen, Mixtral-Instruct, and DeepSeek.}
\label{fig:mechanism}
\end{figure*}

\subsection{Native Scores Identify Local Contributions}

At a fixed hidden state, the local output of prefix $1{:}j$ is
\begin{equation}
 y^{1{:}j}_{l,t}=\sum_{i=1}^{j}w_iE_{e_i}(h_{l,t}).
\end{equation}
The relative contribution of slot $j$ is
\begin{equation}
    d_{l,t,j}=\frac{\|y^{1{:}j}_{l,t}-y^{1{:}j-1}_{l,t}\|_2}
    {\|y^{1{:}j}_{l,t}\|_2+\epsilon}.
    \label{eq:local-damage}
\end{equation}
This statistic measures the current layer's output change under removal of the slot. Across the Qwen C4 and WikiText control samples, the prefix-relative score has mean Spearman correlation 0.865 with $d$. Partial correlations controlling raw probability, layer, and slot are 0.634 and 0.637.

Across families, Spearman correlations are 0.862 for Qwen, 0.980 for Mixtral-Instruct, and 0.802 for DeepSeek. A contribution-ranked prefix reference starts from each mandatory first slot and greedily adds the available next slot with the largest measured $d$ until it exhausts the budget. If $\mathcal S_B$ and $\mathcal S_B^\star$ are the additional calls retained by a policy and this matched-size reference, their contribution capture is
\begin{equation}
    \mathrm{Cap}(B)=\frac{\sum_{u\in\mathcal S_B}d_u}
    {\sum_{u\in\mathcal S_B^\star}d_u}.
    \label{eq:capture}
\end{equation}
CED captures 92.5\% to 97.0\% of reference contribution on Qwen, 99.5\% to 99.9\% on Mixtral-Instruct, and 93.2\% to 96.0\% on DeepSeek. It exceeds fixed prefixes, raw-probability thresholds, and random prefixes at every evaluated budget. Router order therefore already encodes most of the local contribution, and the prefix-relative score extracts it more faithfully than raw probability, fixed depth, or random allocation. Figure~\ref{fig:mechanism} summarizes the score controls and contribution capture, and Appendix~\ref{app:score} gives each budget.

\subsection{Executing the Calibrated Budget}

Grouped dispatch turns fewer expert assignments into higher prefill throughput. With Qwen's C4-calibrated $B=3.2$ policy, batch 16 gives 1.44$\times$ prefill throughput at 512 input tokens and 1.39$\times$ at 2,048 tokens, relative to full-budget execution through the same grouped operators. Batch 4 at 2,048 tokens gives 1.42$\times$. The saving carries over to an expert-loop implementation, which gains 1.32$\times$. Appendix~\ref{app:systems} reports absolute latencies.

\paragraph{Integration with a serving engine.}
We integrate CED into vLLM's tensor-parallel expert dispatcher. Skipped slots receive an expert index outside the active range, so native routing builds only the retained expert groups. The engine's grouped expert operators consume these groups, followed by weighted unpermutation, while model weights, attention, scheduling, and KV-cache handling stay unchanged. On an instrumented workload, CED reduces the rows entering grouped expert matrix multiplications from 219,648 to 85,200 per tensor-parallel worker, a 61.2\% reduction and 4.0\% fewer rows than NAEE at the same target. For a batch of 32 requests with 1,024-token prompts and one output token each, CED completes the batch in 0.98\,s, compared with 1.14\,s at full depth on the same dispatcher, a 1.16$\times$ throughput gain.

\section{CED as a Web Serving Control}

CED turns a released MoE checkpoint into a family of service tiers that share one set of weights. Each tier is a single scalar threshold shared across layers and slots (Table~\ref{tab:calibration-thresholds}). Switching tiers changes no weights, caches, or kernels, so a service can move between budgets from one batch to the next. In our vLLM integration, a worker-extension call installs a new threshold without reloading the model, and the native full-depth endpoint remains one call away. Operators can move traffic between quality and cost targets as load changes and return to native execution at any time. Each token's depth depends only on its own router scores and the installed threshold, so a request executes the same experts whichever requests share its batch. Service tiers therefore stay predictable under dynamic batching.

The threshold also bounds the worst case before any request arrives. Equation~\eqref{eq:depth-ceiling} caps every token-layer state at four of Qwen's eight native experts under the $B=3.2$ profile, so the routed expert work of every token stays within half of native execution. Table~\ref{tab:service-tiers} lists the ceiling and realized saving of each Qwen3 and DeepSeek tier. The tightest Qwen tier removes 75.4\% of routed calls with at most three experts per token.

\begin{table}[t]
\caption{One calibration yields service tiers with a guaranteed per-token ceiling. Each tier is a single threshold. Filled cells mark the ceiling of Eq.~\eqref{eq:depth-ceiling} among the $k$ native experts, and $E$ and Saving are measured on held-out C4.}
\label{tab:service-tiers}
\centering
\small
\setlength{\tabcolsep}{4pt}
\begin{tabular}{@{}lrrlrl@{}}
\toprule
& \multicolumn{2}{c}{\textbf{Control}} & \multicolumn{1}{c}{\textbf{Guarantee}} & \multicolumn{2}{c}{\textbf{Held-out C4}} \\
\cmidrule(lr){2-3}\cmidrule(lr){4-4}\cmidrule(l){5-6}
Model & $B$ & $\theta_B$ & Ceiling & $E$ & Saving \\
\midrule
Qwen3 & 2.0 & 0.319 & \ceilingcells{3}{8} 3 & 1.97 & \valuebar{16.588}{75.4\%}{cedgreen} \\
\textcolor{black!55}{$k=8$} & 3.2 & 0.203 & \ceilingcells{4}{8} 4 & 3.15 & \valuebar{13.332}{60.6\%}{cedgreen} \\
 & 4.5 & 0.136 & \ceilingcells{7}{8} 7 & 4.45 & \valuebar{9.768}{44.4\%}{cedgreen} \\
\midrule
DeepSeek & 2.0 & 0.294 & \ceilingcells{3}{6} 3 & 1.99 & \valuebar{14.718}{66.9\%}{cedgreen} \\
\textcolor{black!55}{$k=6$} & 3.0 & 0.186 & \ceilingcells{5}{6} 5 & 2.98 & \valuebar{11.066}{50.3\%}{cedgreen} \\
 & 4.0 & 0.130 & \ceilingcells{6}{6} 6 & 3.98 & \valuebar{7.392}{33.6\%}{cedgreen} \\
\bottomrule
\end{tabular}
\end{table}

The throughput benefit concentrates where Web services spend their prefill compute. Retrieval-augmented answering and agent observation steps feed long contexts into the model \citep{lewis2020rag,yao2023react}, and disaggregated serving dedicates instances to prefill \citep{zhong2024distserve,patel2024splitwise}. Our measured batches with 8,192 to 32,768 prompt tokens reach 1.39$\times$ to 1.44$\times$ prefill throughput, providing an expert-compute control for this tier on the same hardware.

\section{Related Work}

\paragraph{Learned expert activation.}
Conditional expert computation scales capacity under bounded per-token work \citep{jacobs1991adaptive,shazeer2017outrageously,lepikhin2020gshard,fedus2022switch,du2022glam}. Learned routing policies adjust active expert counts, introduce inexpensive experts, or allocate computation by input difficulty \citep{zeng2024adamoe,jin2024moepp,huang2024harder,yue2024adak,guo2024dynmoe}. Expert-choice routing lets tokens receive variable numbers of experts \citep{zhou2022expertchoice}, and LongCat-Flash adds zero-computation experts to a released model \citep{longcat2025flash}. Adaptive computation more broadly halts or exits early per input \citep{graves2016act,schuster2022calm}. These methods train the routing behavior, whereas CED operates on the released checkpoint's existing router and expert weights.

\paragraph{Skipping with frozen models.}
Lu et al.'s NAEE skips Mixtral's second expert when its routing weight falls below a layer-specific multiple of the leading weight \citep{lu2024experts}. Each multiplier is the median ratio on calibration data, and its released implementation and top-$k$ tail-mass extension both renormalize the retained weights. Mixture Compressor combines a related ratio rule with protection for salient tokens \citep{huang2025mcmoe}. AdapMoE uses Fisher-weighted sensitivity and a shared threshold to choose between one and two experts \citep{zhong2024adapmoe}. ACE estimates expert contributions from frozen parameters, maps skipping budgets to thresholds with one unlabeled pass, and renormalizes the retained gates \citep{xu2026ace}. Concurrent work likewise finds that renormalizing over the retained experts distorts the expert gain set in training, and normalizes by a larger expert set instead \citep{chen2026halving}. CED keeps the native coefficients that NAEE and ACE renormalize and pairs them with prefix-relative scores and one average-budget threshold.

\paragraph{Budget allocation and compression.}
Layer-level allocation complements CED. LExI derives static layerwise expert counts from model weights under a total budget \citep{chittyvenkata2025lexi}, and Alloc-MoE sets layer budgets from sensitivity profiles and fills them with the highest routing scores among co-processed tokens \citep{liu2026allocmoe}. CED instead sets each token's depth from its own scores, and per-layer CED thresholds can realize an allocator's layer targets. MoDES combines layer importance with modality-specific thresholding for multimodal skipping \citep{huang2026modes}. Expert merging and compression shrink the checkpoint itself \citep{li2023mcsmoe,chen2024hcsmoe}, whereas CED keeps every expert and serves every tier from one checkpoint.

\paragraph{Expert execution and serving.}
Tutel, MegaBlocks, and DeepSpeed-MoE optimize sparse expert execution \citep{hwang2022tutel,gale2022megablocks,rajbhandari2022deepspeedmoe}. Orca, vLLM, and SGLang raise serving throughput through iteration-level scheduling, paged memory, and prefix sharing \citep{yu2022orca,kwon2023vllm,zheng2024sglang}. Sarathi-Serve chunks prefills \citep{agrawal2024sarathi}, and DistServe and Splitwise separate prefill from decoding \citep{zhong2024distserve,patel2024splitwise}. For Web services, SCOOT tunes engine configurations for service-level objectives and BATON re-batches inference requests \citep{cheng2025scoot,baton2025www}. ExpertFlow coordinates expert caching and scheduling \citep{he2024expertflow}, and Lynx uses batch-aware expert selection \citep{gupta2024lynx}. These systems tune how a fixed model is served, and CED removes expert computation that they leave untouched.

\section{Conclusion}

CED-MoE turns frozen expert routers into a calibrated compute control. A shared threshold allocates native prefixes across tokens and layers, and unchanged mixture coefficients preserve retained expert contributions. Across three MoE families, this combination lowers perplexity relative to NAEE in every calibrated comparison and relative to ACE on Qwen at matched depth, with the largest margins at the tightest budgets, and native coefficients supply most of the gain. On full GSM8K, CED retains 96.7\% of full-budget accuracy with 60.8\% fewer routed expert calls and leads NAEE by 16.9 points, and a C4-calibrated threshold holds its target depth during reasoning. Grouped execution converts the saving into up to 1.44$\times$ prefill throughput, and the same policy runs inside vLLM. The threshold also fixes a per-token expert ceiling, so serving systems obtain both an average operating point and a maximum routed depth. Under every selection rule we evaluate, native coefficients matter more than which experts are skipped. CED gives released MoE checkpoints a reversible family of quality-compute operating points for Web inference services. Adopting it needs one threshold per tier, with no retraining and no new expert kernels.

Because calibration reads only router traces, a new MoE release obtains a complete set of service tiers from one unlabeled pass, without retraining, labels, per-layer search, or checkpoint surgery. The thresholds run inside an existing serving engine beneath its batching and memory management, which makes expert depth a deployable control for Web-scale inference and for the assistants and agents built on it.
\label{sec:main-end}

\label{sec:references}
\bibliographystyle{ACM-Reference-Format}
\bibliography{ced_moe_refs}

\appendix
\section{Experimental Protocols}
\label{app:implementation}

The implementation uses Transformers 4.57.6 \citep{wolf2020transformers}, PyTorch 2.10.0, SDPA attention, and BF16 weights. Consecutive layer groups reside on separate accelerators, and full and reduced budgets share the grouped expert path.

Qwen3 has 48 MoE blocks, 128 experts per block, and $k=8$. Mixtral has 32 blocks, eight experts, and $k=2$. DeepSeek has 59 MoE blocks, 160 routed experts, $k=6$, two always-active shared experts, and a dense first block.

\paragraph{Inputs and uncertainty.}
C4 training and validation documents are disjoint. Deterministic offsets select 513-token windows, giving 512 inputs and targets. WikiText-2 is concatenated in corpus order into disjoint 513-token blocks. Policies share token IDs. PPL exponentiates the mean window NLL. Bootstrap units are documents or blocks for PPL and questions for generation.

\paragraph{Native execution at full depth.}
Full-depth grouped execution matches native Transformers perplexity within 0.1\% on both corpora for all three families (Tables~\ref{tab:quality-qwen}, \ref{tab:quality-mixtral}, and~\ref{tab:quality-deepseek}). Restoring every native slot recovers the native expert sum exactly, and the BF16 implementations agree to this precision.

\subsection{Full-Test Generation}
\label{app:generation}

Five fixed training examples demonstrate reasoning ending in \texttt{\#\#\#\#} and a numerical answer. Native chat formatting disables thinking. Greedy decoding uses left-padding masks, valid positions, and KV caching, stopping at EOS or 2,048 tokens. Counts exclude padding and post-EOS positions. Native Transformers uses batch one across eight disjoint shards, and grouped policies use batch 16 on identical prompts.

Strict scoring reads the number after the delimiter. Numerical scoring, our primary metric, then checks a boxed answer, an explicit answer phrase, and the last number in that order. Table~\ref{tab:generation-details} reports both scores. Strict scoring costs CED only 0.07 points but NAEE 14.9 points, since many NAEE answers break the requested format.

\begin{table}[t]
\caption{Full GSM8K test results. Strict requires the requested delimiter. Numerical uses the common answer matcher, with bars on a 0 to 100\% scale. Tokens is mean generated length including EOS. Cap counts responses reaching 2,048 tokens.}
\label{tab:generation-details}
\centering
\small
\setlength{\tabcolsep}{4pt}
\begin{tabular}{@{}lrrrr@{}}
\toprule
& \multicolumn{2}{c}{\textbf{Accuracy (\%)$\uparrow$}} & \multicolumn{2}{c}{\textbf{Generation}} \\
\cmidrule(lr){2-3}\cmidrule(l){4-5}
Policy & Strict & Numerical & Tokens & Cap \\
\midrule
Native HF & 93.78 & \valuebar{20.632}{93.78}{cedblue} & 159.0 & 0 \\
Full grouped & 94.54 & \valuebar{20.799}{94.54}{cedblue} & 159.0 & 0 \\
\midrule
\rowcolor{cedwin}
\textbf{CED} & 91.36 & \valuebar{20.115}{\textbf{91.43}}{cedgreen} & 163.4 & 2 \\
Raw $p_j$ & 88.48 & \valuebar{19.498}{88.63}{cedblue} & 155.8 & 0 \\
NAEE layer & 59.59 & \valuebar{16.396}{74.53}{cedblue} & 199.0 & 10 \\
\bottomrule
\end{tabular}
\end{table}

Table~\ref{tab:gsm-paired} pairs CED with NAEE and raw-probability thresholds on identical prompts.

\begin{table}[t]
\caption{CED against NAEE and raw-probability thresholds on all 1,319 GSM8K questions. $\Delta$ is the numerical-accuracy gain of CED in points with a 95\% question-bootstrap interval. W/L counts questions that only CED or only the baseline answers correctly. $p$ is the exact McNemar test. NAEE uses layer-specific thresholds.}
\label{tab:gsm-paired}
\centering
\small
\setlength{\tabcolsep}{4pt}
\begin{tabular}{@{}lrrr@{}}
\toprule
Baseline & $\Delta$ [95\% CI] & W/L & $p$ \\
\midrule
NAEE & \textbf{+16.9} [14.6, 19.3] & 252/29 & $1.5\times10^{-45}$ \\
Raw $p_j$ & \textbf{+2.8} [1.2, 4.4] & 75/38 & $6.4\times10^{-4}$ \\
\bottomrule
\end{tabular}
\end{table}

\section{Complete Language-Model Comparisons}
\label{app:quality}

Tables~\ref{tab:quality-qwen}, \ref{tab:quality-mixtral}, and~\ref{tab:quality-deepseek} report all CED and NAEE profiles from frozen full-depth C4 calibration. Layer-specific NAEE quantiles target equal depth in each layer. The original Mixtral rule skips expert two below the layer's median $p_2/p_1$. NAEE and CED renorm.\ redistribute the original selected mass as in Eq.~\eqref{eq:renormalization}, and ACE runs its released implementation \citep{xu2026ace}. ACE always keeps the top-1 expert, truncates the cumulative mass of its contribution scores, and renormalizes the retained gates. Its global threshold is fitted on the same C4 traces.

\begin{table}[t]
\caption{Qwen3-30B-A3B with frozen C4 thresholds. $E$: achieved depth. Per corpus: 128 windows and 65,536 loss tokens.}
\label{tab:quality-qwen}
\centering
\small
\setlength{\tabcolsep}{4pt}
\begin{tabular}{@{}lrrrr@{}}
\toprule
& \multicolumn{2}{c}{\textbf{C4 validation}} & \multicolumn{2}{c}{\textbf{WikiText-2 test}} \\
\cmidrule(lr){2-3}\cmidrule(l){4-5}
Policy / target $B$ & $E$ & PPL$\downarrow$ & $E$ & PPL$\downarrow$ \\
\midrule
Native HF & 8.000 & 14.81 & 8.000 & 11.66 \\
Full grouped & 8.000 & 14.80 & 8.000 & 11.65 \\
\addlinespace[0.6pt]
\rowcolor{cedwin}
\textbf{CED 2.0} & 1.969 & \textbf{25.79} & 1.962 & \textbf{20.92} \\
NAEE global 2.0 & 2.055 & 69.78 & 2.052 & 84.83 \\
NAEE layer 2.0 & 2.053 & 69.19 & 2.050 & 85.59 \\
\addlinespace[0.6pt]
\rowcolor{cedwin}
\textbf{CED 3.2} & 3.149 & \textbf{18.90} & 3.115 & \textbf{14.73} \\
NAEE global 3.2 & 3.277 & 21.86 & 3.254 & 19.28 \\
NAEE layer 3.2 & 3.276 & 22.07 & 3.252 & 19.64 \\
ACE 3.2 & 3.272 & 21.93 & 3.244 & 19.11 \\
\addlinespace[0.6pt]
\rowcolor{cedwin}
\textbf{CED 4.5} & 4.452 & \textbf{16.49} & 4.399 & \textbf{12.92} \\
NAEE global 4.5 & 4.538 & 16.84 & 4.508 & 13.65 \\
NAEE layer 4.5 & 4.540 & 16.88 & 4.507 & 13.79 \\
\addlinespace[0.6pt]
CED renorm. 3.2 & 3.245 & 21.49 & 3.215 & 19.04 \\
\bottomrule
\end{tabular}
\end{table}

\begin{table}[t]
\caption{Mixtral-8x7B-v0.1 with frozen C4 thresholds. $E$: achieved depth. Per corpus: 128 windows and 65,536 loss tokens.}
\label{tab:quality-mixtral}
\centering
\small
\setlength{\tabcolsep}{4pt}
\begin{tabular}{@{}lrrrr@{}}
\toprule
& \multicolumn{2}{c}{\textbf{C4 validation}} & \multicolumn{2}{c}{\textbf{WikiText-2 test}} \\
\cmidrule(lr){2-3}\cmidrule(l){4-5}
Policy / target $B$ & $E$ & PPL$\downarrow$ & $E$ & PPL$\downarrow$ \\
\midrule
Native HF & 2.000 & 7.52 & 2.000 & 5.69 \\
Full grouped & 2.000 & 7.52 & 2.000 & 5.69 \\
\addlinespace[0.6pt]
\rowcolor{cedwin}
\textbf{CED 1.2} & 1.183 & \textbf{7.90} & 1.174 & \textbf{6.12} \\
NAEE global 1.2 & 1.196 & 8.20 & 1.185 & 6.42 \\
NAEE layer 1.2 & 1.197 & 8.20 & 1.186 & 6.43 \\
\addlinespace[0.6pt]
\rowcolor{cedwin}
\textbf{CED 1.5} & 1.474 & \textbf{7.69} & 1.442 & \textbf{5.86} \\
NAEE global 1.5 & 1.489 & 7.78 & 1.455 & 6.00 \\
NAEE layer 1.5 & 1.489 & 7.79 & 1.455 & 6.02 \\
\addlinespace[0.6pt]
\rowcolor{cedwin}
\textbf{CED 1.8} & 1.785 & \textbf{7.56} & 1.740 & \textbf{5.74} \\
NAEE global 1.8 & 1.786 & 7.59 & 1.742 & 5.78 \\
NAEE layer 1.8 & 1.785 & 7.60 & 1.737 & 5.78 \\
\addlinespace[0.6pt]
CED renorm. 1.5 & 1.489 & 7.78 & 1.455 & 6.00 \\
NAEE median & 1.489 & 7.79 & 1.455 & 6.01 \\
\bottomrule
\end{tabular}
\end{table}

\begin{table}[t]
\caption{DeepSeek-V2 with frozen C4 thresholds. $E$: achieved depth. Per corpus: 128 windows and 65,536 loss tokens.}
\label{tab:quality-deepseek}
\centering
\small
\setlength{\tabcolsep}{4pt}
\begin{tabular}{@{}lrrrr@{}}
\toprule
& \multicolumn{2}{c}{\textbf{C4 validation}} & \multicolumn{2}{c}{\textbf{WikiText-2 test}} \\
\cmidrule(lr){2-3}\cmidrule(l){4-5}
Policy / target $B$ & $E$ & PPL$\downarrow$ & $E$ & PPL$\downarrow$ \\
\midrule
Native HF & 6.000 & 7.81 & 6.000 & 6.07 \\
Full grouped & 6.000 & 7.81 & 6.000 & 6.06 \\
\addlinespace[0.6pt]
\rowcolor{cedwin}
\textbf{CED 2.0} & 1.989 & \textbf{8.29} & 1.978 & \textbf{6.74} \\
NAEE global 2.0 & 2.024 & 9.91 & 2.000 & 8.85 \\
NAEE layer 2.0 & 2.023 & 9.85 & 2.001 & 8.75 \\
\addlinespace[0.6pt]
\rowcolor{cedwin}
\textbf{CED 3.0} & 2.985 & \textbf{7.98} & 2.955 & \textbf{6.34} \\
NAEE global 3.0 & 3.017 & 8.50 & 2.972 & 7.10 \\
NAEE layer 3.0 & 3.017 & 8.47 & 2.971 & 7.08 \\
\addlinespace[0.6pt]
\rowcolor{cedwin}
\textbf{CED 4.0} & 3.983 & \textbf{7.89} & 3.915 & \textbf{6.18} \\
NAEE global 4.0 & 4.005 & 8.05 & 3.947 & 6.46 \\
NAEE layer 4.0 & 4.005 & 8.04 & 3.947 & 6.45 \\
\addlinespace[0.6pt]
CED renorm. 3.0 & 3.020 & 8.45 & 2.988 & 7.03 \\
\bottomrule
\end{tabular}
\end{table}

\subsection{Matching Executed Counts}

The count-matched comparison anchors the budget to CED's realized depth on the C4 evaluation inputs. A single multiplicative factor rescales the comparator's global threshold or all its per-layer thresholds. Repeated count-only passes bracket this factor and bisect until achieved depth is within 0.005 of the anchor. Perplexity is evaluated after fixing the factor. Table~\ref{tab:matched-depth} reports exact achieved depths.

\section{Calibration and Prefix Properties}
\label{app:calibration}

Table~\ref{tab:calibration-thresholds} gives the shared CED thresholds and achieved calibration depths. On the frozen calibration traces, all profiles reach the requested depth within 0.00031 experts. The Qwen, Mixtral, and DeepSeek traces contain 3,145,728, 2,097,152, and 3,866,624 routed token-layer records, respectively. Each family needs one full-depth pass over 65,536 calibration tokens, and every budget profile reuses these traces.

\begin{table}[t]
\caption{CED thresholds shared across layers and extra slots, fitted on 128 C4 training windows per model. Bars use $[0,1/2]$. Trace depth estimates retained calls on full-depth traces. Values have six decimals.}
\label{tab:calibration-thresholds}
\centering
\small
\setlength{\tabcolsep}{4pt}
\begin{tabular}{@{}lrrr@{}}
\toprule
Model & Target $B$ & $\theta_B$ & Trace depth \\
\midrule
Qwen3 & 2.0 & \valuebar{14.041}{0.319121}{cedblue} & 2.000046 \\
 & 3.2 & \valuebar{8.943}{0.203247}{cedblue} & 3.199994 \\
 & 4.5 & \valuebar{5.993}{0.136203}{cedblue} & 4.500000 \\
\midrule
Mixtral & 1.2 & \valuebar{19.264}{0.437824}{cedblue} & 1.199695 \\
 & 1.5 & \valuebar{14.684}{0.333720}{cedblue} & 1.499777 \\
 & 1.8 & \valuebar{9.417}{0.214034}{cedblue} & 1.799927 \\
\midrule
DeepSeek & 2.0 & \valuebar{12.958}{0.294489}{cedblue} & 2.000000 \\
 & 3.0 & \valuebar{8.194}{0.186228}{cedblue} & 3.000000 \\
 & 4.0 & \valuebar{5.721}{0.130014}{cedblue} & 4.000000 \\
\bottomrule
\end{tabular}
\end{table}

\begin{figure}[t]
\centering
\includegraphics[width=\columnwidth]{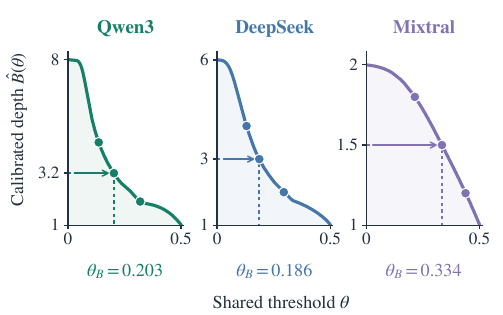}
\caption{Calibrated budget curves from one pass of unlabeled C4 training traces. Markers are the nine calibrated profiles. Every curve is monotone, starts at the native depth $k$, and reaches one expert at $\theta=1/2$.}
\Description{Three decreasing curves of calibrated depth against the shared threshold for Qwen3, DeepSeek, and Mixtral, with three calibrated profile markers on each curve.}
\label{fig:budget-curve}
\end{figure}

Figure~\ref{fig:budget-curve} plots the complete budget curve $\hat B(\theta)$ of each family from one pass of C4 training traces, so every requested budget is a point on a monotone curve.

\paragraph{Layer allocation.}
At the middle profiles, the shared CED threshold gives layer means from 2.88 to 3.49 experts on Qwen, 1.31 to 1.65 on Mixtral, and 2.79 to 4.68 on DeepSeek. Layer-specific NAEE quantiles stay within 3.18 to 3.33, 1.44 to 1.51, and 2.99 to 3.05 experts, respectively (Figure~\ref{fig:layer-depth}).

\paragraph{Monotone scores.}
For descending nonnegative scores and $P_j=\sum_{i\le j}p_i$, the extra-slot scores satisfy $p_{j+1}/(P_{j+1}+\epsilon)\le p_j/(P_{j+1}+\epsilon)\le p_j/(P_j+\epsilon)$, so thresholding them yields a prefix. With $\epsilon=0$, a common positive rescaling of all probabilities leaves every ratio unchanged.

\paragraph{Empirical budget resolution.}
A tied group of $g$ effective scores moves the average depth by $g/C$. Choosing the nearer adjacent threshold therefore incurs at most $g/(2C)$ budget error, plus at most $1/(2C)$ from rounding the extra-call count, and score ties stay together.

\section{Local Contributions}
\label{app:score}

The local analysis uses 1.2M sampled token-layer-slot records per family. Qwen traces come from C4 validation, Mixtral-Instruct traces from WikiText-2 test, and DeepSeek traces from WikiText-2 validation and test. All use length-512 sequences. The Qwen score-control panels use 50,000 records from each of C4 and WikiText validation. Table~\ref{tab:oracle-grid} gives contribution capture at each budget. Fixed prefixes use a score-independent mixture of adjacent depths where the target is fractional. Random rows average five prefix-allocation seeds.

\begin{table}[htbp]
\caption{Local contribution capture (\%) relative to the contribution-ranked prefix at equal budget. Higher is better. Shading tracks capture from 65 to 100\%. CED leads every row. Random averages five allocations. Mixtral-I is Mixtral-8x7B-Instruct.}
\label{tab:oracle-grid}
\centering
\small
\setlength{\tabcolsep}{4pt}
\begin{tabular}{@{}lrrrrr@{}}
\toprule
Model & $B$ & Random & Fixed & Raw $p_j$ & \textbf{CED} \\
\midrule
Qwen & 2.0 & \cellcolor{cedgreen!5!white}70.6 & \cellcolor{cedgreen!22!white}90.6 & \cellcolor{cedgreen!13!white}79.8 & \cellcolor{cedgreen!24!white}\textbf{92.5} \\
 & 3.2 & \cellcolor{cedgreen!8!white}74.7 & \cellcolor{cedgreen!24!white}92.8 & \cellcolor{cedgreen!20!white}88.3 & \cellcolor{cedgreen!26!white}\textbf{95.9} \\
 & 4.5 & \cellcolor{cedgreen!13!white}80.3 & \cellcolor{cedgreen!25!white}94.4 & \cellcolor{cedgreen!24!white}92.7 & \cellcolor{cedgreen!27!white}\textbf{97.0} \\
\addlinespace[2pt]
Mixtral-I & 1.5 & \cellcolor{cedgreen!4!white}69.1 & \cellcolor{cedgreen!2!white}67.9 & \cellcolor{cedgreen!26!white}95.4 & \cellcolor{cedgreen!30!white}\textbf{99.5} \\
 & 1.8 & \cellcolor{cedgreen!17!white}84.8 & \cellcolor{cedgreen!16!white}84.1 & \cellcolor{cedgreen!30!white}99.6 & \cellcolor{cedgreen!30!white}\textbf{99.9} \\
\addlinespace[2pt]
DeepSeek & 3.0 & \cellcolor{cedgreen!4!white}69.4 & \cellcolor{cedgreen!19!white}87.0 & \cellcolor{cedgreen!13!white}80.3 & \cellcolor{cedgreen!24!white}\textbf{93.2} \\
 & 4.0 & \cellcolor{cedgreen!11!white}78.0 & \cellcolor{cedgreen!22!white}90.5 & \cellcolor{cedgreen!18!white}85.9 & \cellcolor{cedgreen!27!white}\textbf{96.0} \\
\bottomrule
\end{tabular}
\end{table}

The slot-only control ranks earlier slots first. The selected-slot statistic in Figure~\ref{fig:mechanism}b takes the highest-scoring fraction $(B-1)/(k-1)$ of later-slot records. With $\mathcal R$ denoting all records and $\mathcal T_B$ the selected subset, it is
\begin{equation}
 L(B)=\frac{|\mathcal T_B|^{-1}\sum_{u\in\mathcal T_B}d_u}
 {|\mathcal R|^{-1}\sum_{u\in\mathcal R}d_u}.
\end{equation}
On C4, CED reaches 1.73 [1.72, 1.75], raw probability reaches 1.60 [1.58, 1.62], and the slot-only score reaches 1.68 [1.66, 1.70]. On WikiText the values are 1.75 [1.73, 1.76], 1.60 [1.58, 1.61], and 1.68 [1.67, 1.70]. Brackets give bootstrap 95\% intervals.

\section{Execution Measurements}
\label{app:systems}

Table~\ref{tab:latency-details} gives absolute batched prefill latency. Full-budget and CED runs alternate, with two warmups and nine measured repetitions per policy, and routing counts come from separate untimed passes.

\begin{table}[t]
\caption{Qwen batched prefill latency (ms) through the same grouped path. CED uses the C4-calibrated $B=3.2$ threshold. Values are medians of nine interleaved repetitions. Bars span 1 to 1.5$\times$ speedup.}
\label{tab:latency-details}
\centering
\small
\setlength{\tabcolsep}{4pt}
\begin{tabular}{@{}rrrrr@{}}
\toprule
Batch & Input & Full & CED & Speedup$\uparrow$ \\
\midrule
4 & 512 & 226.7 & 212.3 & \valuebar{2.986}{\textbf{1.07}$\times$}{cedgreen} \\
 & 2048 & 647.6 & 457.3 & \valuebar{18.314}{\textbf{1.42}$\times$}{cedgreen} \\
\addlinespace[2pt]
16 & 512 & 621.4 & 430.4 & \valuebar{19.529}{\textbf{1.44}$\times$}{cedgreen} \\
 & 2048 & 2731.9 & 1960.4 & \valuebar{17.315}{\textbf{1.39}$\times$}{cedgreen} \\
\bottomrule
\end{tabular}
\end{table}

An expert-loop implementation at batch four and input length 512 takes 3,940.6\,ms at full depth and 2,975.7\,ms under CED, a 1.32$\times$ prefill speedup.

\subsection{vLLM Integration}

The engine experiment uses vLLM 0.22.1 with two-way tensor parallelism, unquantized BF16 weights, eager execution, a 4,096-token model limit, a 32,768-token scheduling limit, at most 32 concurrent requests, and no prefix caching. The batch measurement uses fixed C4 prompts, one warmup, and the median of three measured repetitions.

The integration thresholds native top-$k$ weights, assigns skipped slots an expert index outside the active expert range, and passes the original hidden states to native routing. The operator constructs per-expert counts for the retained assignments, and grouped expert matrix multiplications consume these counts. The final unpermutation applies zero weights to skipped slots. Tier switching uses the vLLM worker-extension mechanism. One collective call installs the new threshold on every tensor-parallel worker, and the next batch dispatches with it while weights and the KV cache stay in place.

On the fixed diagnostic workload, full depth places 219,648 rows in grouped expert matrix multiplications, CED places 85,200, and shared-threshold NAEE places 88,778, so CED executes 4.0\% fewer rows than NAEE at the same target.

\end{document}